\documentclass[11pt]{article}

 \usepackage{amsmath}
\usepackage[latin1]{inputenc}
\usepackage[usenames,dvipsnames]{color}
\usepackage{graphicx}
\usepackage{amssymb}
\usepackage{verbatim}
\usepackage[usenames,dvipsnames]{color}
\usepackage{mathdots}

\makeatother

  \providecommand{\corollaryname}{Corollary}
  \providecommand{\definitionname}{Definition}
  \providecommand{\lemmaname}{Lemma}
\providecommand{\theoremname}{Theorem}

\newtheorem{prop}{Proposition}

\global\long\def\vecspan{\operatorname{span}}
\global\long\def\argmin{\operatorname*{arg\, min}}
\global\long\def\argmax{\operatorname*{arg\, max}}
\global\long\def\smax{\operatorname*{smax}}
\global\long\def\sign{\operatorname{sign}}
\global\long\def\diag{\operatorname{diag}}
\global\long\def\trace{\operatorname{Tr}}
\newcommand{\RR}{\mathbb{R}}

\newcommand{\trans}{^{\scriptscriptstyle \top}}
\vspace{3cm}

\usepackage{bbm}
\usepackage{makeidx}

\makeatother

\renewcommand\S{{\bf S}}
\newcommand{\beq}{\begin{equation}}
\newcommand{\eeq}{\end{equation}}

\begin{document}
\begin{center}
\huge{A Regularization Approach for Prediction of  Edges and Node Features in Dynamic Graphs}
\footnote{\\
Emile Richard : ENS Cachan - UniverSud - CMLA, France, 1000mercis, France, emile.richard@cmla.ens-cachan.fr\\
Andreas Argyriou : TTI Chicago, argyriou@ttic.edu\\
Theodoros Evgeniou : Technology Management and Decision Sciences, INSEAD 
Bd de Constance, Fontainebleau 77300, France, theodoros.evgeniou@insead.edu \\
Nicolas Vayatis : ENS Cachan - UniverSud - CMLA,  UMR CNRS 8536 France {nicolas.vayatis@cmla.ens-cachan.fr}

}

\end{center}


\begin{abstract}

  We consider the two problems of predicting links in a dynamic graph
  sequence and predicting functions defined at each node of the graph. 
 In many applications, the solution of one problem is useful for solving the other. 
Indeed, if these functions reflect node features, then they are related through the
  graph structure. In this paper, we formulate a hybrid
  approach that simultaneously learns the structure of the graph and predicts 
  the values of the node-related functions. Our approach is based on the optimization of a
  joint regularization objective. We empirically test the benefits of
  the proposed method with both synthetic and real data. The results
  indicate that joint regularization improves prediction performance
  over the graph evolution and the node features.
\end{abstract}

\section{Introduction}

Forecasting the behavior of systems with multiple responses has been a
challenging problem in the context of many applications such as
collaborative filtering, financial markets, or bioinformatics, where
responses may be, respectively, movie ratings, stock prices, or
activity of genes within a cell. Statistical modeling techniques have
been widely applied for learning {\em multivariate time series}
either in the multiple linear regression setting \cite{Breiman97} or
with autoregressive models \cite{Tsay05}. More recently, kernel-based
regularized methods have been developed for multitask learning
\cite{Evgeniou05, Argyriou07}.  These approaches share in common the
use of the correlation structure between input variables to enhance
prediction of every single output. Frequently, the correlation structure is
assumed to be given or is estimated separately. A discrete encoding
of correlations between variables can be modeled as a graph so that
learning the dependence structure amounts to performing graph
inference through the discovery of unobserved edges on the graph. The
latter problem is interesting {\em per se} and is known as
{\em link prediction}, where it is assumed that only part of the
graph is actually observed \cite{Kleinberg07, Kolar11}. This situation
occurs in various applications such as recommender systems, social
networks, or proteomics, and the appropriate tools can be found among
matrix completion techniques \cite{Srebro05, Candes09, Abernethy09}.
In the realistic setup of a time-evolving graph, matrix completion was
also used and adapted to take into account the dynamics of the
features of the graph \cite{Richard10}.

In this paper, our goal is to simultaneously predict multiple outputs
defined over the vertices of a time-evolving graph and learn
the structure of the graph. 
One important assumption we make is that the network effect is a cause
and a symptom at the same time. Consider, for instance, the problem of
forecasting sales based on purchase data from an e-commerce market
where the semantic information is not reliable. The data can be
represented as the (bipartite) graph of users and products connected
by purchases which evolves over time according to the purchase
history. We expect that within a cluster of co-purchased items, the
sales should be correlated, and reversely, correlated sales should
induce more edges between corresponding items. A similar situation
arises in the context of financial data where the graph reflects
dependencies between stocks and one aims at both predicting stock
prices and inferring the dependence structure.  We build on the
approach proposed in \cite{Chen10} which shows that, in the case where
graph structure information is available about the dependencies
between input variables, the efficiency of multivariate predictions
can be enhanced. In this paper, we tackle a problem of broader scope,
that is, we assume that the underlying graph is unknown. We propose a
formulation of the problem as a regularized risk minimization which
balances both objectives of prediction and matrix completion at the
same time. The proposed formulation is convex with respect to each
target variable (predictor and adjacency matrix) separately but not
jointly convex. We study the performance of this joint optimization
approach through a set of experiments on real and synthetic data. We mention that a byproduct of our investigations is the introduction of a generative model using latent factors for creating artificial data sets.
We discuss empirical convergence, implementation issues and performance 
with respect to each of the two objectives (multivariate prediction and graph inference).

The problem of simultaneous prediction of multivariate time series
with graph dependence structure and inference of the graph structure
is presented in Section 2. We also present the joint regularization
approach for solving this problem. In Section 3 we discuss the algorithm used and
the computation of relevant features within this regularization approach.
Numerical experiments are discussed and results on synthetic and real
data sets are presented in Section 4.


\section{Learning Dynamic Graph Features and Edges}
\label{sec:setup}

\subsection{Setup}

We first introduce notations for the main objects of interest in the paper:
\begin{itemize}
\item
 a sequence of $T$ undirected weighted graphs
with $n$ vertices and their corresponding $n \times n$ adjacency matrices $A_t\in\S^{n}_{\geq 0}, t \in
\{1,2,...,T\}$, where $\S^n_{\geq0}$ denotes the set of $n\times n$ symmetric matrices with 
nonnegative entries. 
We assume that, the weights of the
edges are nondecreasing functions of time, which in the unweighted case means edges do not disappear but new ones may appear over time.

\item a multivariate time series of node features $\bigl(X_t\bigr)_{1\le t \le T}$ so that $\forall t, X_t \in \RR^{n \times q}$. 
 We denote by $X_t^{(i)} \in \RR^q$ the $i$th row of $X_t$ representing the feature vector of node $i$ at time $t$. We assume that this series depends on
the series of graphs via some function $\omega:\RR^{n\times n}\to
\RR^{n\times q}$, that is, $X_t = \omega(A_t)$. 

\end{itemize}

Given the values of $(A_t, X_t)$, for $t
\in \{1,2,...,T\}$, the main goal is to simultaneously predict the future value
$X_{T+1}$ and the future adjacency matrix
$A_{T+1}$.  To this end, we introduce:
\begin{itemize}
\item A descriptor matrix $\Phi_t\in \RR^{n \times d}$, $t\in \{1,2,...,T\}$, which encodes the
  past information contained both in the time series $(X_s)_{1 \le s
    \le t}$ and in the sequence of adjacency matrices $(A_s)_{1 \le s
    \le t}$ into a row vector of dimension $d$ for each node.
\item A matrix-valued prediction function $f: \RR^{n \times d}\to\RR^{n\times
    q}$ which relates the variable $X_{t+1}$ to the past
  information $\Phi_t$ so that  $X_{t+1 }$ is close to $f(\Phi_t)$. The
  function $f$ is unknown and has to be estimated from past data. We
  denote the $i$-th row component of $f$ with the notation $f^{(i)}$, $i\in \{1,2,...,n\}$.
  The $n$ prediction functions $f^{(i)}$ are assumed to belong to the same Hilbert
  space $\mathcal{H}$, equipped with the norm $\|\cdot
  \|_{\mathcal{H}}$. The norm of $f \in \mathcal{H}^n$ is defined by
  $\|f\|_{\mathcal{H}^n} := \sqrt{ \sum_{i=1}^n\|f^{(i)}\|^2_{\mathcal{H}}}$.
\item An unknown matrix $S\in\S^n_{\geq0}$ whose elements indicate
how likely it is that there is a nonzero value at the corresponding position of
matrix $A_{T+1}$. The most likely edges at time $T+1$ are the ones 
corresponding to the largest values in $S$.
\end{itemize}
\subsection{Motivation}
The challenge in the prediction problem on real graphs evolving over
time is to relate global and slowly varying node features with local and
sudden changes of edges. Thus it is reasonable to assume that the evolution of
the graph is governed by unobserved {\em latent factors} \cite{Sarkar05,Sarkar10}, denoted by $\Psi_t$, which {\em evolve
smoothly}. For example, in social networks or marketing applications, latent factors
could be psychological factors which account for the choices of consumers. The effect of such factors
on the graph can then be measured using the adjacency matrix. For
instance, the clustering coefficient captures homophilic behavior, the
node degree is a good indicator of popularity and the pagerank
refers to centrality and influence, the trends can be read in the evolution of node degrees and the tastes are believed to be reflected on the first singular vectors of the adjacency matrix. 
A possible formulation of the latent factor model
sets $\Psi_t$ to be a pair of matrices $(U_t,V_t)$ governing the structure of $A_t$ through $A_t = U_t V_t\trans$, where the slow evolution of latent factors is guided by some unobserved underlying mechanism. 
A rigorous framework would involve a statistical model for the noise and a specific form of dependence of the matrices
on their recent history. A formal example will be provided in Section 4.
Both the latent factors and their evolution being unobservable, we assume that some relevant and observable features $X_t$ partly capture the information contained in the latent factors. 



\subsection{Assumptions}
The approach studied in this paper relies on
the assumption that exploiting simultaneously the {\em structure of the
graph} and the {\em dynamics of the time series} should improve both
prediction and graph learning. Given the information collected over
the period $t=1, \ldots, T$, which is contained in the feature matrices
$\Phi_t$, we want to learn $f$ and $S$ which satisfy the following
assumptions:
\begin{itemize}
\item {\bf [A1] Low rank.} $S$ has low rank. This is a standard
  assumption in matrix completion problems \cite{Srebro05, Candes09}.
  The rationale is that the factors $U_t, V_t$ (see Sec. 2.2) should be of small dimension. 

\item {\bf [A2] Graph growth regularity.} The growth of the graph
  exhibits regularity over time, that is, $S$ should be close to the
  last adjacency matrix $A_T$. We will use the Frobenius distance between these
  two matrices as an estimate of the error of the learning process 
  with respect to graph inference.
\item {\bf [A3] Feature growth regularity.} The features of
  the predicted graph are assumed to be close to the predicted features. As a consequence,
  we assume that the norm of $f(\Phi_T) - \omega(S)$ should be kept small.
\item {\bf [A4] Stationarity.} The dependence mechanism between
  $X_{t+1}$ and $\Phi_t$ can be inferred. Therefore, it is reasonable
  to assume stationarity of the joint distribution of $(X_{t+1},
  \Phi_t)$.


\item {\bf [A5] Regularity of the predictors over the graph.}
  Neighboring vertices $i$ and $j$ in the graph should correspond to similar
  prediction functions $f^{(i)}$ and $f^{(j)}$.
\end{itemize}

\subsection{Formulation of the Optimization Problem}
We now introduce a regularization based formulation which 
reflects each of the previous assumptions. To this end, we introduce
additional notation. For any matrix $M$, 
$\text{Tr}(M)$ denotes the trace of $M$ and  
$\| M\|_\text{F} :=\sqrt{ \text{Tr} (M\trans M)}$ 
denotes the Frobenius norm of $M$.
We also define $ \| M \|_* := \sum_{k=1}^n \sigma_k(M)$ ,
the nuclear norm of a matrix $M$, where
$\sigma_k(M)$ denotes the $k$-th largest singular value of $M$. We
recall that a singular value of matrix $M$ corresponds to the square
root of an eigenvalue of $M\trans M$ and that the
nuclear norm is a convex surrogate of the rank. We also define the matrix
$\Delta(f) := \left( \|f^{(i)} - f^{(j)}\|_\mathcal{H}^2 \right)_{i,j=1}^n$.
We introduce a convex loss function $\ell :\RR^{n\times q} \times \RR^{n\times q} \to \RR_+$ which measures prediction
errors. Given the past history  $\{ (X_t, A_t) ~:~  1 \le t \le T \}$, the proposed optimization problem is then to minimize over
$(f,S) \in \mathcal{H}\times \S^n_{\geq0}$ the following functional:
\begin{equation}\label{eq:optim}
  \mathcal{L}(f, S)  := \sum_{t=1}^{T-1} \ell(f(\Phi_t),X_{t+1}) + 
\frac{\kappa}{2} \|f\|^2_{\mathcal{H}^n}  + \ell(f(\Phi_T), \omega(S)) $$ 
$$+ \tau 
\| S\|_* +   \frac{\nu}{2}  \| S  -  A_T  \|_\text{F}^2 +\lambda
\trace \bigl( S\trans \Delta(f)\bigr) \,,
\end{equation}
where $\lambda$, $\tau$ and $\nu$ are positive regularization
parameters. Typical choices of the loss function $\ell$ are the 
square loss or the sum of hinge loss errors per component. Note that
using different loss functions allows one to adapt the method to different
problems, such as regression or classification.

We now describe each of the separate
terms in the above formulation.
\begin{itemize}
\item The first term, $J_1(f) = \sum_{t=1}^{T-1}
  \ell(f(\Phi_t), X_{t+1})$ $+\frac{ \kappa}{2} \|f\|_{\mathcal{H}^n}$,
  corresponds to the prediction task as in standard regularized risk
  minimization in RKHS for supervised learning. 
\item The second term, $J_2(S) = \mu\| S\|_* + \frac{1}{2} \| S -
  A_T \|_\text{F}^2$, with $\mu = \tau/\nu$, represents the low-rank
  matrix denoising objective (static link prediction) and reflects assumptions A1 and A2.
\item The third term, $J_3(f,S) =  \ell(f (\Phi_T), \omega(S))$, 
  penalizes the difference between the predicted
  features and the features of the predicted matrix. This term reflects
  assumption A3. 
\item The last term, $J_4(f, S) = \trace \bigl( S\trans \Delta(f)\bigr)$
relates the contributions of $f$ and $S$ based on assumption A.
\end{itemize}

We point out the connection of work on matrix completion \cite{Candes08,Negahban11, Koltchinskii11} to the minimization of $\ell(f(\Phi_T), \omega(S)) + \mu \|S\|_*$.
Also, in \cite{Richard10}, the authors use the minimizer $\hat{f}$ of $J_1$ and solve the subproblem $\argmin_S J_2(S) + J_3(S,\hat{f})$.

\subsection{Linear Models}
For simplicity of presentation, we 
now consider the case of $\mathcal{H}$ being an RKHS corresponding to a linear kernel. 
In addition, each $f^{(i)}$ is a linear function represented by a $d \times
q$ matrix $W^{(i)}$, so that $f^{(i)}(\Phi_t)=\Phi^{(i)}_t W^{(i)} $. Then
we may assume $\|f^{(i)}\|_{\mathcal{H}} = \|W^{(i)}\|_F$ and the graph regularity
term can be expressed in terms of the graph Laplacian of $S$. 
Recall that the Laplacian is defined as the operator $\Lambda$ such that 
$\Lambda(S) = D-S$ where $D = \diag(d_1, \ldots, d_n)$
and $d_i = \sum_{j=1}^n S_{ij}$ are the degrees of the graph.  
A standard computation gives the following expression for $J_4$:
\[
\trace \bigl( S\trans \Delta(f)\bigr) = \sum_{1 \le i,j \le n} S_{ij}
\|W^{(i)} - W^{(j)}\|_F^2 
\]
where $W = (W^{(1)}, \ldots, W^{(n)}) \in \mathbb{R}^{n \times d
  \times q}$. We use the standard extension of the Frobenius norm to
3-tensors, $\|W\|_F^2 = \sqrt{ \sum_{i=1}^n\|W^{(i)}\|^2_F}$
and the notations:
\begin{align*}
Q(W,\Lambda, V) & := \sum_{1 \le i,j \le n} \Lambda_{ij} 
\trace({W^{(i)}}\trans V^{(j)})~,\\ 
\Delta(W) & := \left( \|W^{(i)} - W^{(j)}\|_F^2 \right)_{i,j=1}^n~.
\end{align*}
We also assume in this work that the node features are linear
functions of adjacency matrices : $\omega(A_t) = A_t\Omega$ for some
$\Omega \in \RR^{n \times q}$. Note that degree, inter/intra cluster
degrees and projection onto specific linear subspaces are such
features, but not clustering coefficients or statistics on path
lengths. We later detail how we chose $\Omega$ in our experiments.


\medskip

\noindent {\em Working with a known and static graph.} This problem
was considered in \cite{Chen10} showing interestingly that feature
prediction can be improved when using graph structure in the
regularized optimization formulation. In their work, the graph penalty
(our $J_4$) takes the form:
\[
\sum_{1 \le i,j \le n} S_{ij} \|W^{(i)} - W^{(j)}\|_1~,
\]
where $S$ is fixed and we can see that an $L_1$ norm is used for this
penalty term instead of $L_2$ in our case. Moreover, their approach
allows negative interactions between vertices. This indicates that
there are clearly many different variants to explore.


\section{Learning Algorithm}
\label{sec:opt}

In this section we address several issues related to the optimization 
problem \eqref{eq:optim} and discuss the learning algorithm we will use.
The algorithm is a standard {\em projected gradient} method,
which projects on the constraint set  $\mathcal{E}$ defined below. The main
challenges are the nonconvexity of $J_4$ and the nondifferentiability of $J_2$
in the objective.

\subsection{Convexity}
In the experiments conducted, we
have learned linear functions $f^{(i)}$. In this case, the term
$ \sum_{1 \le i,j \le n} S_{ij} \|W^{(i)} - W^{(i)}\|_F^2$ is convex
with respect to one of the variables, if the other is fixed, but not jointly convex with
respect to $(W,S)$. In order to ensure
convergence of an optimization algorithm to a desired minimum, we
determine a domain where $\mathcal{L}$ is convex. We use intuition from what happens with a similar optimization problem in dimension 2. Indeed, consider the minimization of $(w,s) \rightarrow sw^2 + \alpha s^2 + \beta w^2$ which is the simplified functional in the degenerate case where $n=1$. The  eigenvectors of the  Hessian are positive iff $w^2 \leq \alpha \beta$, so this condition defines the convexity region. By adding the quadratic terms in $S$ and $W$ to $Q(W,\Lambda(S),W)$, let us define $$\Psi
(S,W) :=\frac{\kappa}{2} \|W\|_F^2 + \frac{\nu}{2} \|S-A_T\|_F^2 $$
$$+\lambda  Q(W, \Lambda(S), W).$$
If we suppose that the entries of $Z = S-A_T$ are nonnegative, 
then the minimizer of $\Psi$ is the trivial solution $S = A_T, W = 0$.
In fact,
\begin{multline}
\Psi (S,W) = \frac{\kappa}{2} \|W\|_F^2 +\frac{\nu}{2} \|Z\|_F^2 +\\
\lambda\bigg ( Q(W, \Lambda(A_T), W) +  Q(W, \Lambda(Z), W) \bigg )\geq 0
\end{multline}
and for $Z = 0,W=0$, $\Psi (S,W) =0$. 

We prove that $\mathcal{L}$ is convex over a set $ \mathcal{E} $
around the minimizer of $\Psi$ and we will ensure henceforth
that the descent algorithm takes place inside this
convex domain.

\begin{prop} The function $\Psi$ is convex in the interior of the set:
$$ \mathcal{E} = \bigg \{  S\in \S^n_{\geq0},W \in \mathbb{R}^{n \times d\times q} 
~~\bigg | ~~  \|W\|_F \leq \frac{\sqrt{\nu\kappa}}{2\lambda\bigl ( \sqrt{n}+1 \bigr )}  \bigg \} \ .  $$ 

\end{prop}
\noindent {\bf Proof.} 
We introduce the slack variable $Z = S-A_T$ and isolate the quadratic
part of $\Psi(Z_0+Z, W_0 + W)$ for some $(Z_0,W_0)$ : 
\begin{multline}
\nonumber
R(Z,W) :=
\frac{\nu}{2} \|Z\|_F^2 + \frac{\kappa}{2} \|W\|_F^2 +\\
\lambda \Bigl ( 2 Q( W , \Lambda(Z), W_0 ) +
Q(W , \Lambda(A_T+Z_0), W) \Bigr).
\end{multline} 
Thanks to
Cauchy-Schwarz and the basic norm property $\|AB\|_F \leq \|A\|_F
\|B\|_F$,
$$ Q( W , \Lambda(Z), W_0) 
\geq - \|W\|_F \| \Lambda(Z)\|_F \|W_0\|_F. $$
We have $\Lambda(Z) = D-Z$. We get, again by Cauchy-Schwarz 
$$\|D\|_F^2 = \sum_{i=1}^n (\sum_{j=1}^n Z_{i,j})^2 \le  \sum_{i=1}^n n \sum_{j=1}^n Z_{i,j}^2 = n \|Z\|_F^2$$
and therefore 
$$\|\Lambda(Z)\|_F = \|D-Z\|_F \le \|D\|_F + \|Z\|_F \le (\sqrt{n}+1) \|Z\|_F.$$ 
On the other hand $ Q (W , \Lambda(A_T+Z_0),W ) \geq 0$, so
$$R(Z,W) \geq  \frac{\nu}{2} \|Z\|_F^2 + \frac{\kappa}{2} \|W\|_F^2 
-2  \lambda(\sqrt{n}+1)  \|W\|_F \| Z\|_F \|W_0\|_F .$$

Letting $z = \|Z\|_F, w = \|W\|_F, w_0 = \|W_0\|_F$, we have 
$$R(Z,W) \geq  \frac{\nu}{2} z^2 + \frac{\kappa}{2} w^2 - 2   \lambda(\sqrt{n}+1) wzw_0 \ .$$
 $\psi_{w_0} : (z,w) \mapsto \frac{\nu}{2} z^2 + \frac{\kappa}{2} w^2 - 2   \lambda(\sqrt{n}+1) wzw_0$ is a quadratic form.
Therefore $R(Z,W)$ is always nonnegative if $\psi_{w_0}$ is positive semidefinite positive. That is, if $\nu + \kappa \geq 0$
(always true) and $\nu \kappa - 4 \lambda^2\bigl ( \sqrt{n} + 1\bigr )^2 w_0^2 \geq 0$,
and this completes the proof. 


\subsection{Smoothing the Nuclear Norm}
For optimization purposes,
we chose to replace the nonsmooth nuclear norm term by a smooth
approximation, in the spirit of the optimization literature, for example 
\cite{Nesterov05}, $$g_\eta(S) = \max_Z \bigg \{ \langle
S,Z\rangle - \frac{\eta}{2}\|Z\|_F^2 \, | \, \sigma_1(Z) \leq 1 \bigg \}.$$
Each of these parameterized functions is a lower bound that approaches $\|S\|_*$
as $\eta\to 0^+$, while being differentiable. Differentiability is due to
$-g_\eta$ being related to a Moreau envelope \cite{moreau} and specifically
to the squared distance of $\frac{1}{\eta}S$ from the unit ball of the
spectral norm. 
\subsection{Gradient Evaluation}
In the case of a differentiable
loss $\ell$, we explicitly compute $ \nabla\mathcal{L}(W,S)$
as
$$
\frac{\partial \mathcal{L}}{\partial W} (W, S) = \sum_{t=1}^{T-1}
\nabla_W \ell \bigl( W\Phi_t , X_t \bigr) + $$
$$\nabla_W  \ell(W\Phi_T, S \Omega) +\kappa W +  
2\lambda \left(\sum_{j=1}^n\Lambda_{ij}(S)W^{(j)}\right)_{i=1}^n \\$$ 
and
\begin{multline}
\nonumber
\frac{\partial \mathcal{L}}{\partial S} (W, S)  =    \nabla_S  \ell( W\Phi_T, S \Omega)  + 
\lambda  \Delta(W) +\\  \nu \bigl(S- A_T\bigr) + 
\tau U(S) \text{Diag}\left(\min \left(1,\frac{\sigma_i(S)}{\eta}\right)\right)V(S)\trans,
\end{multline}

where $S = U(S) \text{Diag}\bigl(\sigma_1(S),
\ldots, \sigma_n(S) \bigr)V(S)\trans$ is a singular value decomposition of $S$.  
In the case of squared loss, $$ \ell \bigl( W\Phi_t ,X_t \bigr) = \sum_{i=1}^n
\|X_t^{(i)} - \Phi_t^{(i)}W^{(i)}\|_2^2~.$$ 
Therefore:

\begin{align*} 
\nabla_S  \ell \bigl( W\Phi_T ,S \Omega\bigr) & = \left( \bigl( S \Omega \bigr )^{(i)} - \Phi_T^{(i)}W^{(i)} ) \Omega\trans \right )_{i=1}^n ~, \\
\nabla_W  \ell \bigl(W\Phi_t, X_t\bigr) & = \left(\Phi^{(i)\trans}_t( \Phi^{(i)}_t W^{(i)} - X_t^{(i)}  ) \right)_{i=1}^n ~.
\end{align*}


\section{Numerical Experiments}

\subsection{Description of Data Sets}

\subsubsection{Synthetic Data}

\begin{figure}[h]
    \includegraphics[height=5cm]{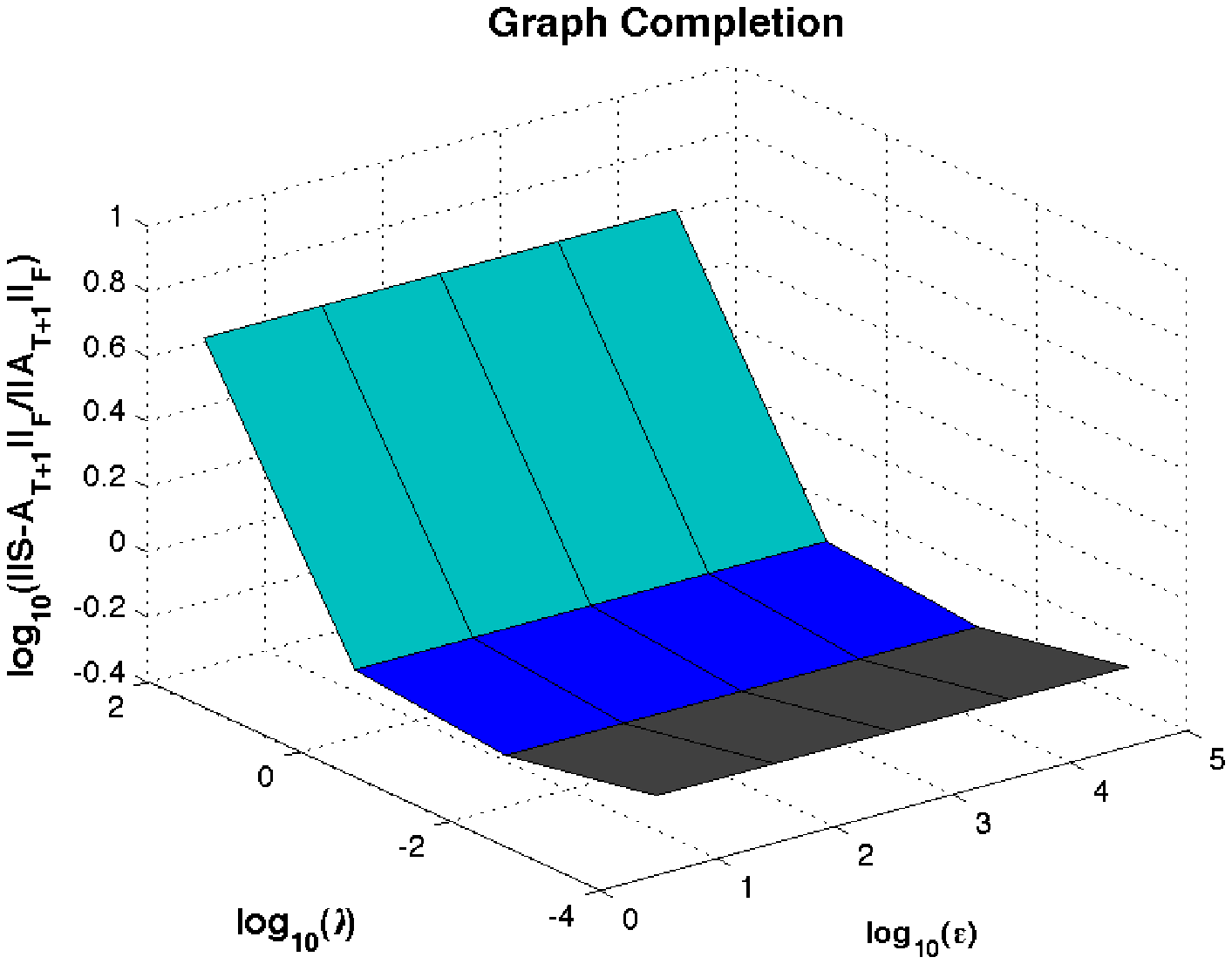}
    \includegraphics[height=5cm]{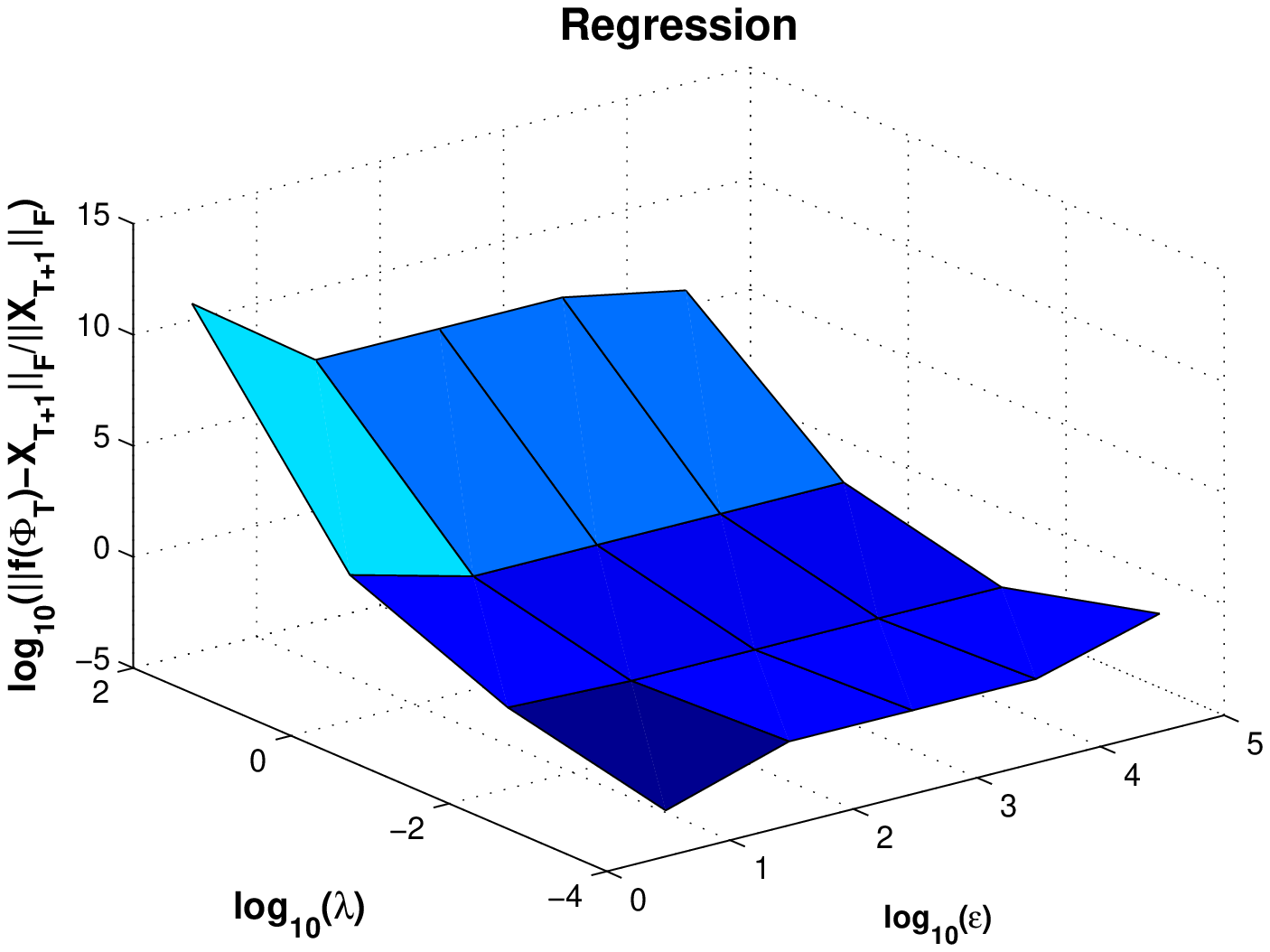}
\caption{Prediction accuracy as a function of $\lambda$ (predictors' regularity on the graph) and $\epsilon$ (nonlinearity of latent factors growth). }
\label{fig:lambda}
\end{figure}

We introduce a generative model for the graph sequence and the node-related features:
$$\left\{
\begin{array}{l}
 X_t =  A_t \Omega\\
 A_t = U_tV_t\trans + z_t  \\
\forall i
\left\{
\begin{array}{ll}
  U_t^{(i)} =U_{t-1}^{(i)} + h(U_{t-1}^{(i)}) + u_{t,i}  \\
    V_t^{(i)} = V_{t-1}^{(i)} + h(V_{t-1}^{(i)}) + v_{t, i} \\
    \end{array}
\right.
    \\
\end{array}
\right.$$ where $ u_{t,i} , v_{t,i}$ are multivariate Gaussian vectors
$\mathcal{N}(0,\delta^2I_r)$ in $\RR^r $, and the entries of $z_t$ are
independently drawn from a centered Gaussian with variance
$\sigma^2$.  We also set
$$h(x) =\epsilon \bigg ( e^{\frac{-\|x-v_1\|^2}{\sigma_1^2}}(x-v_1) +  e^{\frac{-\|x-v_2\|}{\sigma_2}}(x-v_2) \bigg )~,$$ where vectors $v_1,v_2 \in \mathbb{R}_+^r$ are chosen randomly and  $\epsilon,\sigma_1,\sigma_2$ are positive constants.
We use here $n =100$, $r = 4$, $T = 60$ and the entries of $U_0, V_0 \in \mathbb{R}^{n \times r}$ are drawn according to a uniform distribution in $(0,1)$.
Note that such data by construction fulfill the required assumptions, and that the nonlinearity of the smooth function $h$ is what makes the contribution of the Laplacian term $J_4(f,S)$ non-trivial, we highlight this phenomenon in Figure \ref{fig:lambda}.

\subsubsection{Marketing Data}
We use data from several months of book purchase history of a major  e-commerce website to
evaluate our method. We tested our method on the task of predicting the cross-selling and the sales volumes on two types of data sets : (1) the top-sold items and (2) at an aggregated level among different product categories.
\begin{enumerate}
\item {\bf Best-sellers} For our test we selected a set of $300$
  top-sold books, and we aimed at predicting the future sales of those
  books, and also predicting the cross-sales, or co-purchases. We
  consider the set of products being the vertices of the dynamic graph
  sequence, and construct the `co-purchase' graph as follows.  At time
  $t$ one can represent the state of market by a binary $\#users
  \times \# items$ matrix $M_t$ where entry $(i,j)$ is non-zero if
  user $i$ has purchased item $j$ so far. We define the co-purchase
  matrix $A_t = M_t\trans M_t$ as the adjacency matrix of the weighted
  graph. The co-purchase graph consists of the products as nodes
  and the weights of edges are given by the number of co-purchases of
  two products.  We use 32 weeks (8 months) of observation for
  learning, and predict the evolution of sales volumes of the books
  over the next 5 weeks. The target feature, namely the sales volume
  is obtained using the bipartite graph of users and products. The
  degrees of items in the bipartite graph represent the sales volumes.
  Our algorithm outputs simultaneousely predictions of the top-selling
  items and a matrix $S$ predicting the
  cross-sellings. 
\item {\bf Categories } Predicting the sales and co-sales at the aggregated level of product categories is relevant for marketers and supply-chain managers. We collected data from $n=195$ book categories over a history of $T=74$ week periods. The entry $(i,j)$ of $A_t$ counts the number of users having purchased items of category $i$ at time $t-1$ who purchased at least an item from category $j$ at time $t$. The features of interest are the weekly sales of each category and the number of new users in each category. 
\end{enumerate}

\subsection{Features and Descriptors}
We recall that our setup consists of a regression-type model for the prediction of $(X_{T+1}, A_{T+1})$ given the available information at time $T$ which includes the previous adjacency matrices $A_1, \ldots, A_T$ of size $n \times n$ and the node-related feature  matrices $X_1, \ldots, X_T$ of size $n\times q$. Our approach involves a $\Phi_T$ of size $n
\times d$ called {\em descriptor matrix} which encompasses the information contained in the past realizations $(X_t, A_t)$ for $t \le T$. 
\subsubsection{Node Features}
The feature matrix $X_t$ contains two types of node features:
\begin{enumerate}
\item Features whose predicted values are of direct interest. For instance, the volume of sales or popularity of an item, represented by the degree of the related vertex in the purchase graph, or the growth rates of the degree as an indicator of the penetration level \cite{Sood09} of the item in the market.
\item Features which are believed to measure relevant underlying factors that govern graph structure. For instance the clustering coefficient \cite{Watts98} taken as a proxy for homophily is not of direct interest, but is a useful quantity for inferring the future connections of a node.
\end{enumerate}
For simplicity, we set here $X_t = A_t\Omega$ where  $\Omega$ is an $n \times q$ matrix. We assume that the columns of $\Omega$  contain (i) the constant vector ${\bf 1}_n$ (since $A_t {\bf 1}_n$ is the vector of node degrees), (ii) the indicator vectors of several clusters which we  identified in the graph and (iii) the top $k=5$ eigenvectors of $A_t$ which are believed to capture most of the structure of $A_t$. 

\subsubsection{Node Descriptors}
The descriptors  contain the features and also other quantities that are useful for predicting the features. In the present work, we made the following choice:

$$\Phi_t  =\bigl ( A_t\Omega, (A_t - A_{t-1})\Omega, (A_t - 2A_{t-1} + A_{t-2})\Omega \bigr ) $$
$$ = \bigl ( X_t, X_t-X_{t-1}, X_t - 2X_{t-1} + X_{t-2} \bigr ) $$

which is a matrix of size $n \times 3q$ representing features, their velocity and acceleration rates. We remark that time-related statistics of the recent history of $(X_t)_{t\le T}$ (such as moving averages, residual variance), or other quantities  accounting for alternative representations of the time series (such as Fourier or wavelet coefficients, polynomial approximation) could also be included. 

\subsection{Evaluation Metrics}

Various  prediction tasks could be studied in our setup with  specific criteria for optimization and evaluation. We focus here on regression (for feature prediction) and graph completion (for link prediction), but classification of vertices may also be a useful task.

\noindent {\em Regression.} When the effective value of an asset in the future is of interest, a squared error metric is appropriate, leading to the use of $\|X_{T+1}-f(\Phi_T)\|_F$ for evalution and $l(f(\Phi_t),X_t) = \|X_{t+1}-f(\Phi_t)\|_F^2$ in the objective. The reported values are the relative errors $\frac{\|X_{T+1}-f(\Phi_T)\|_F}{\|X_{T+1}\|_F}$.

\noindent {\em Graph Completion.} 
Our method aims  simultaneousely at the prediction of $X_{T+1}$ and $A_{T+1}$. We measure the quality of prediction of $A_{T+1}$ by $\|S-A_{T+1}\|_F$, and the relative values reported are $\frac{\|A_{T+1}-S\|_F}{\|A_{T+1}\|_F}$.

\noindent {\em Classification.} Specific patterns may appear in the time series $X_t$ which we may wish to predict. For instance, predicting top-selling items in a given market is definitely of interest. This problem can be formulated as follows: the matrix $X_t$ represents the sales volumes over a market (each component corresponds to a product), and we assign a $\pm 1$ label depending on the order of  magnitude of the increase of sales volumes over a specific time window.

\subsection{Comparison with Related Methods}

Since the problem which we address (simultaneous graph completion and
feature prediction) is a novel one, there are no direct competing methods to compare with. 
Thus, in our experiments we have considered the following baselines which address the most similar learning problems:
\begin{enumerate}

\item {\em Ridge Regression.} Learning the function $\hat{f} :=
  \argmin_\mathcal{H} J_1$ through the minimization of the penalized
  regression $J_1(f) = \sum_{t=1}^{T-1} \ell(f(\Phi_t), X_{t+1})
  +\frac{ \kappa}{2} \|f\|_{\mathcal{H}^n}^2$ leads us to a prediction
  function $\hat{f}$ which we use for predicting $X_{T+1}$ by
  $f(\Phi_T)$. This approach addresses the prediction of node features
  as independent time series. It does not take the graph sequence into
  account and does not predict the future graph.

\item {\em Rank-free prediction.} We used our formulation with the choice of $\tau=0$. In this case, the loss function ignores the low-rank hypothesis in the graph structure. The prediction process only relies on the dynamics of graph features.

\item {\em Graph completion through shrinkage.} We also considered a method only dedicated to link prediction and matrix denoising, ignoring the feature prediction part of the problem. This method only uses the observation $A_T$ as a noisy version of $A_{T+1}$, and aims at predicting $A_{T+1}$ given only this observation and the low rank hypothesis. We solve the problem $\min_S J_2(S)$ where $J_2(S) = \frac{1}{2}\| S-A_T\|_F^2 + \mu \|S\|_*$ to estimate the adjacency matrix as a low rank approximation of $A_T$. The solution of this problem is obtained by singular value shrinkage \cite{Candes08}, namely if $A_T = U\diag{(\sigma_1(A_T), \ldots, \sigma_n(A_T))}V\trans$ is the singular value decomposition of $A_T$,  then the image of $A_T$ by a shrinkage of parameter $\mu$ is given by $D_\mu(A_T) = U\diag{(\sigma_i(A_T)-\mu)_+}V\trans$ .

\end{enumerate}
\subsection{Results}

\begin{figure}[t]
 \begin{center}
\includegraphics[height=3.4cm]{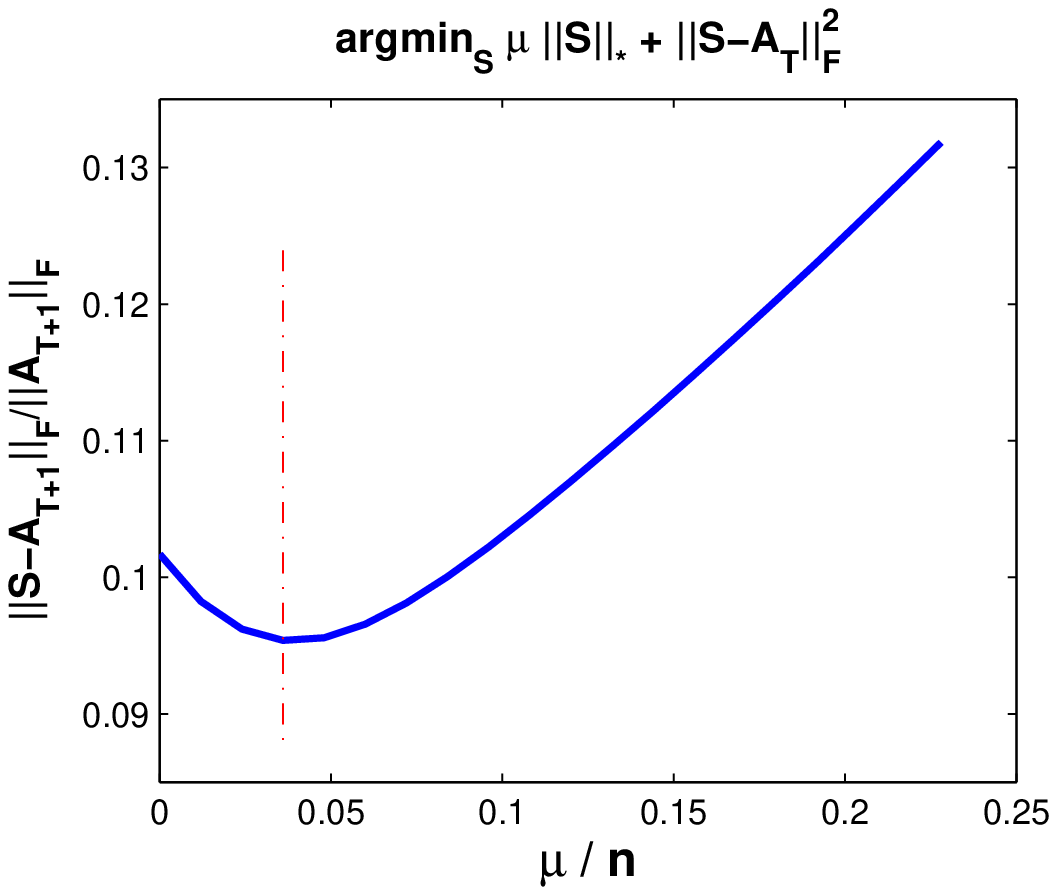}\includegraphics[height=3.4cm]{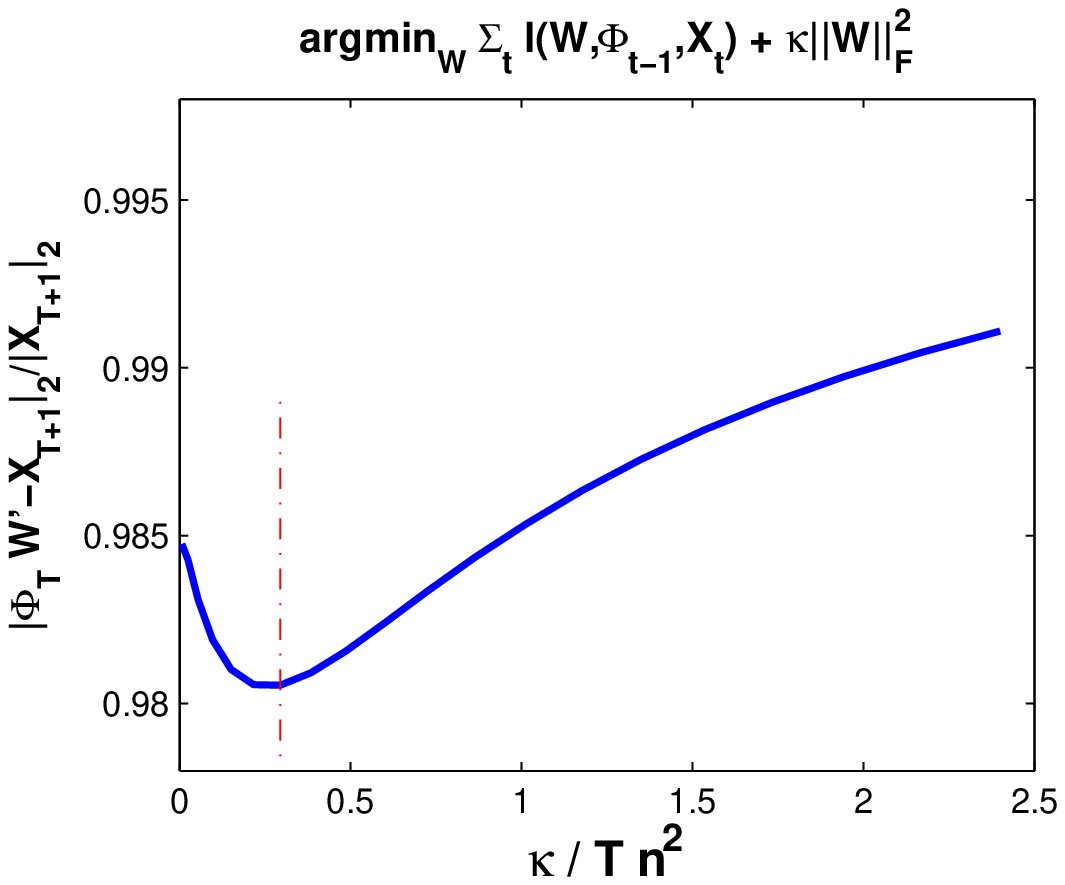}
\includegraphics[height=3.3cm]{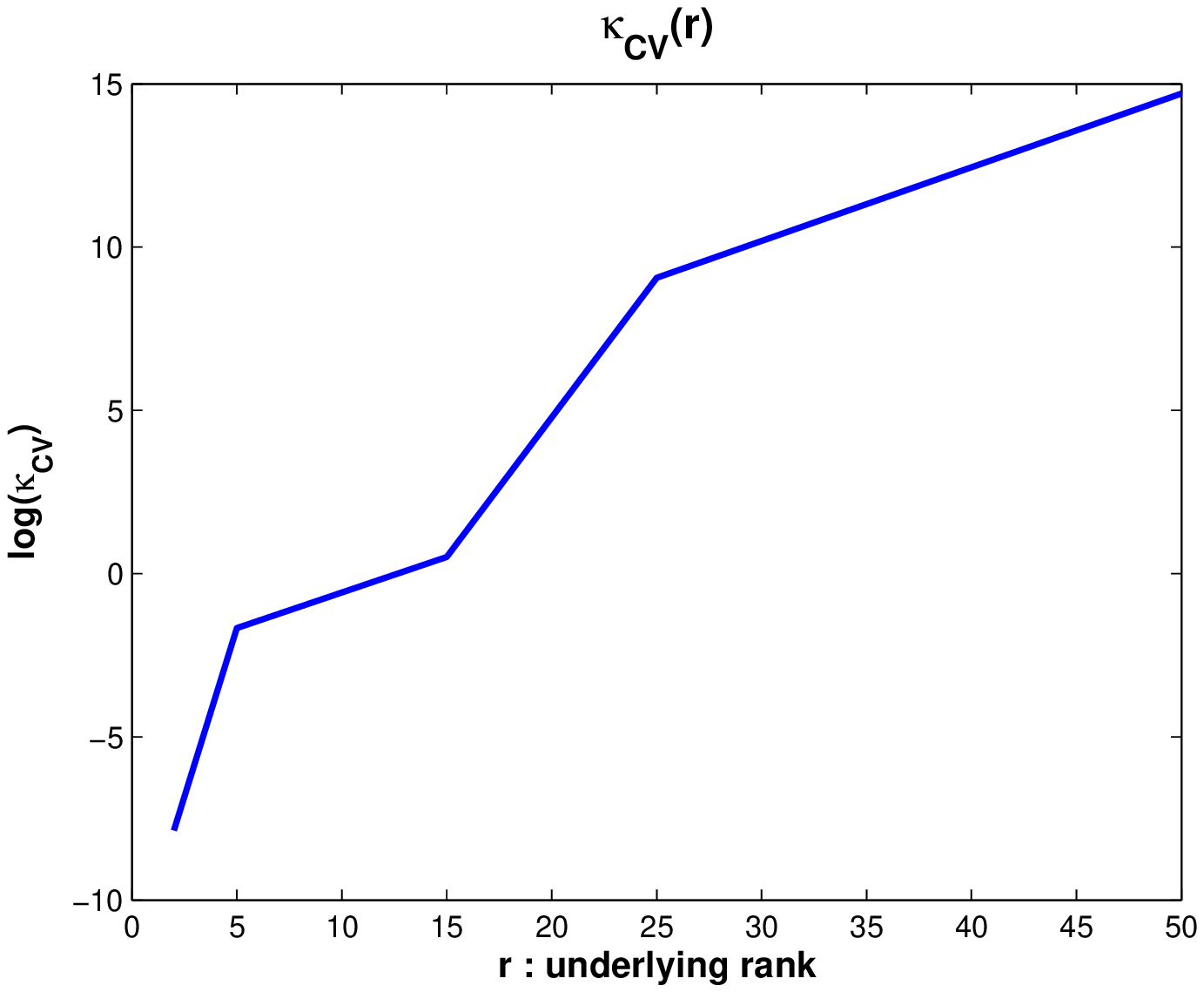}\includegraphics[height=3.3cm]{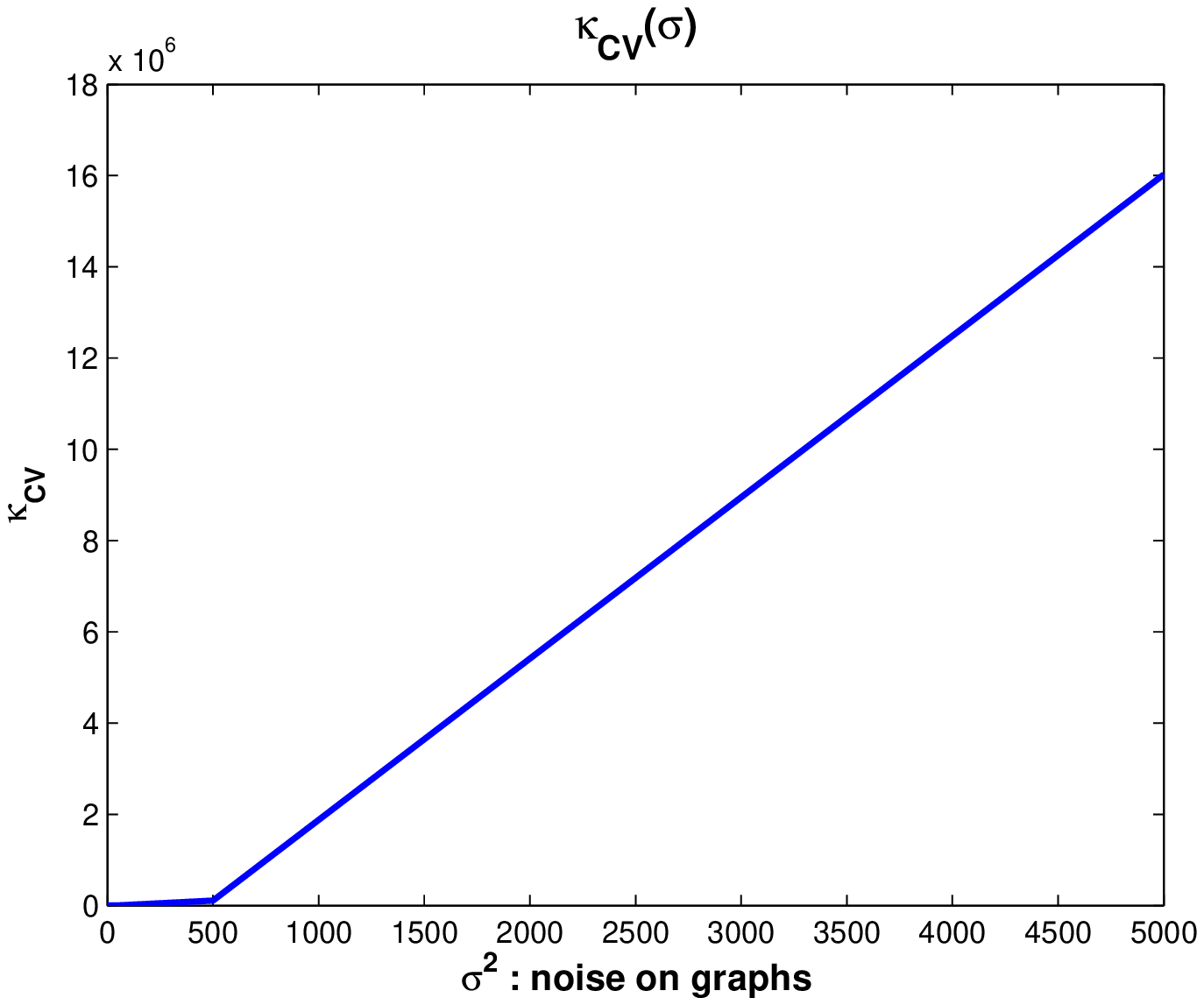}
\includegraphics[height=3.3cm]{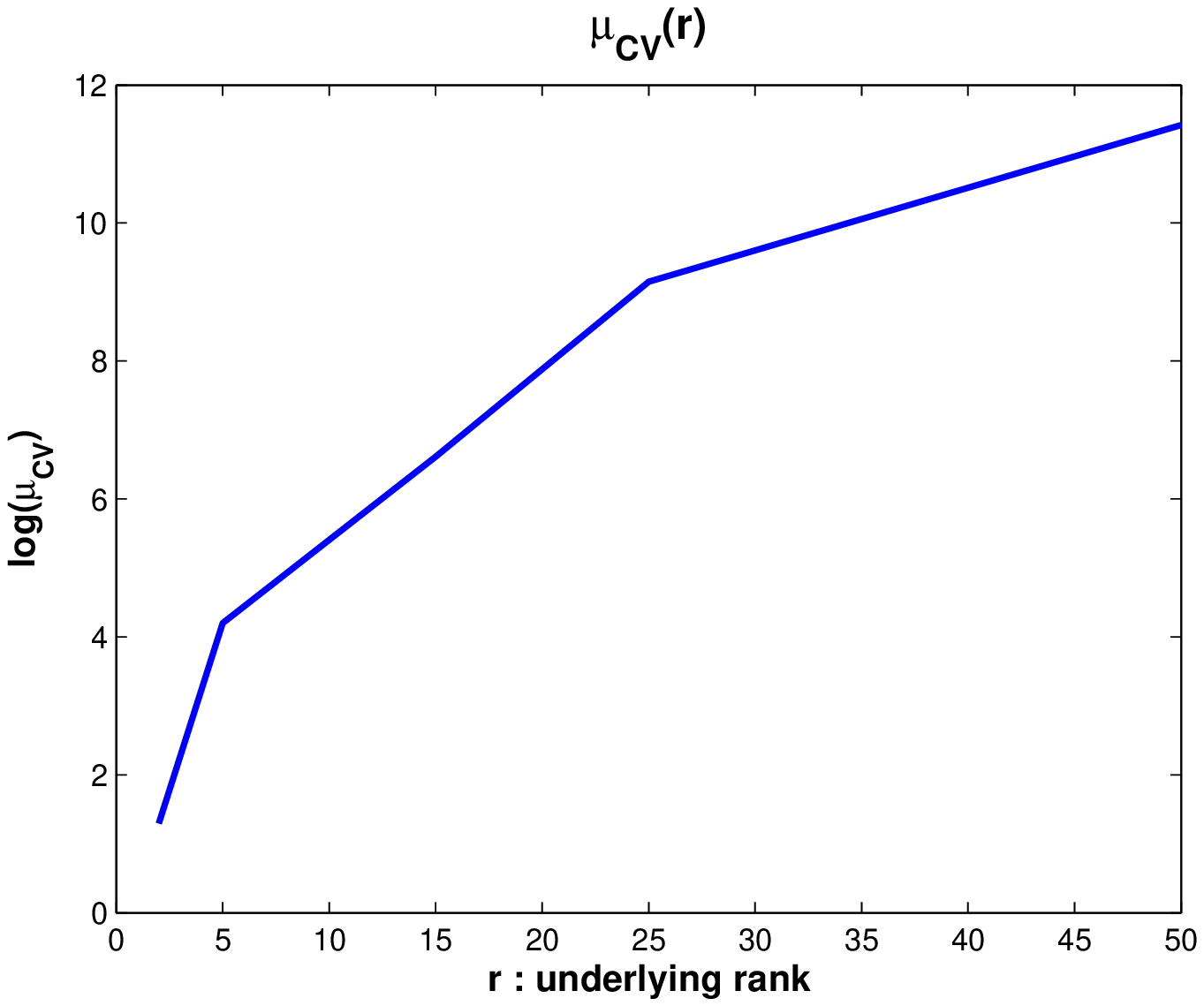}\includegraphics[height=3.3cm]{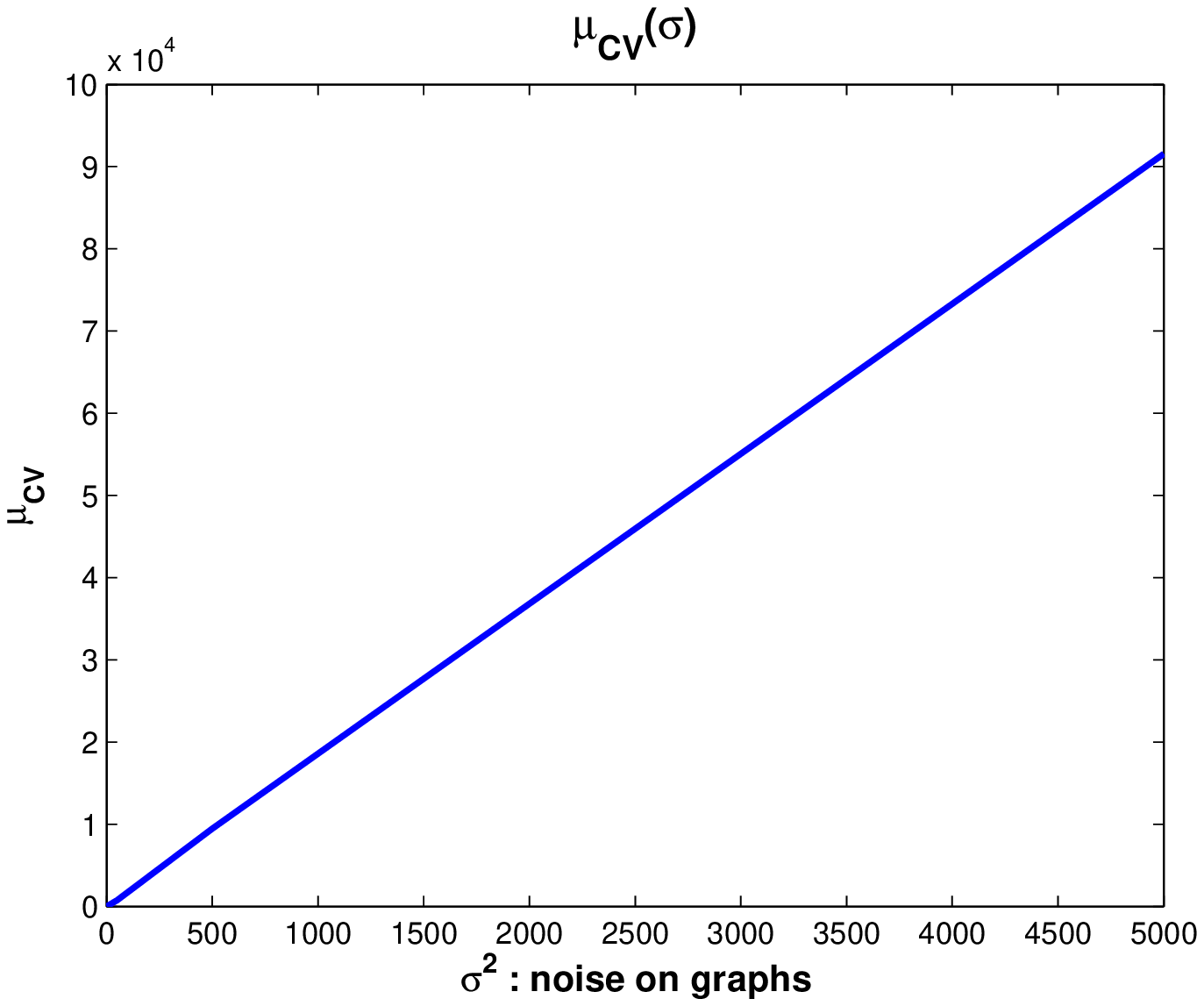}
   \caption{Cross-validation for $\mu$ and $\kappa$ on the two standard objectives of regression and graph completion (matrix denoising) taken separately (top 2 plots) and dependence of constants chosen by cross-validation $\mu_\text{CV},\kappa_\text{CV}$ on generative model constants $r,\sigma$.} 
 \label{fig:cv}
 \end{center}
\end{figure}

\begin{figure}[hbt]

  \includegraphics[height=4.2cm]{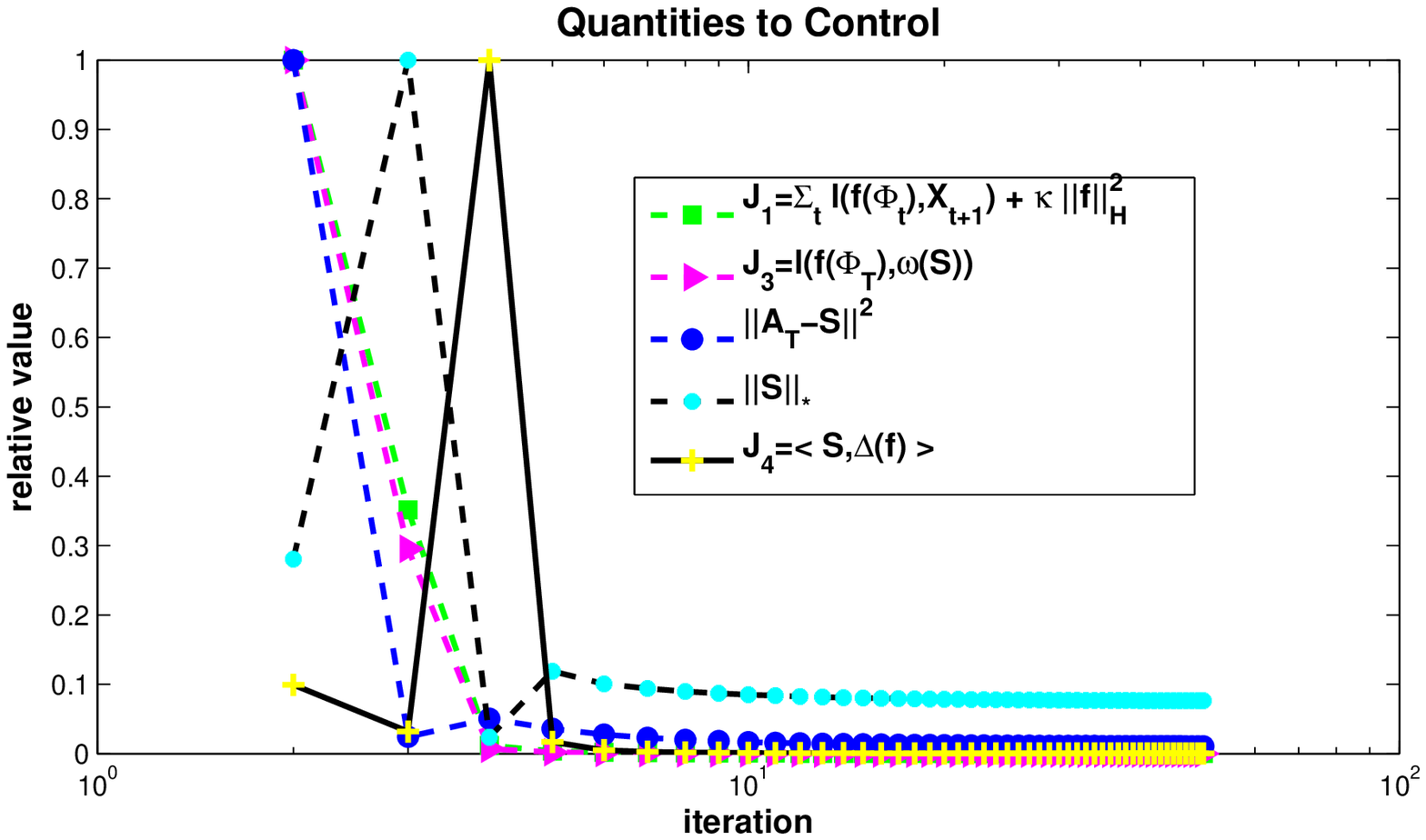}\\
   \includegraphics[height=4.2cm]{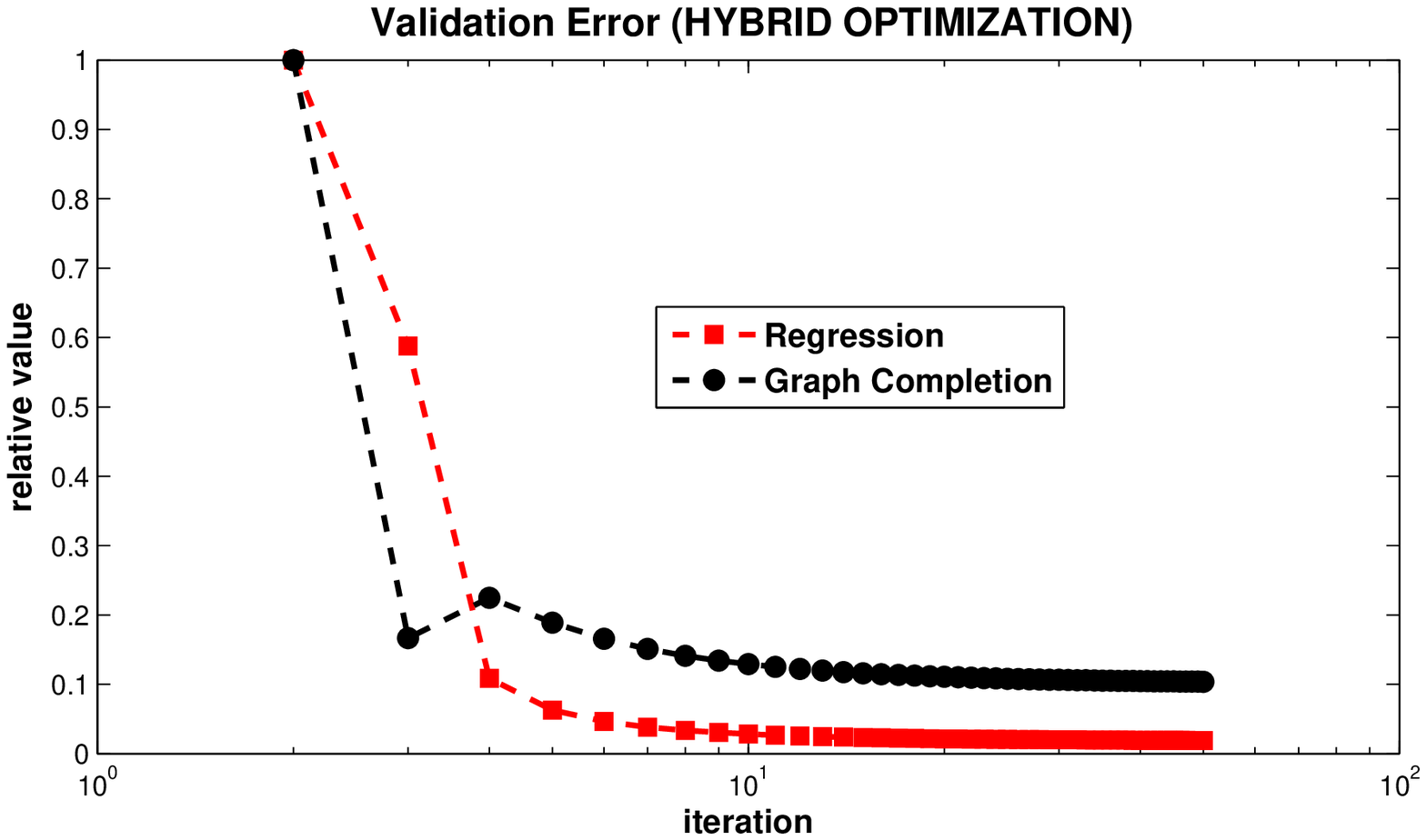}
   \caption{Numerical behavior of different terms of the risk
     functional and validation error versus iteration count of a gradient
     descent algorithm (synthetic data).} \label{fig:desc}

\end{figure}

\begin{figure}[hbt]

    \includegraphics[height=4.5cm]{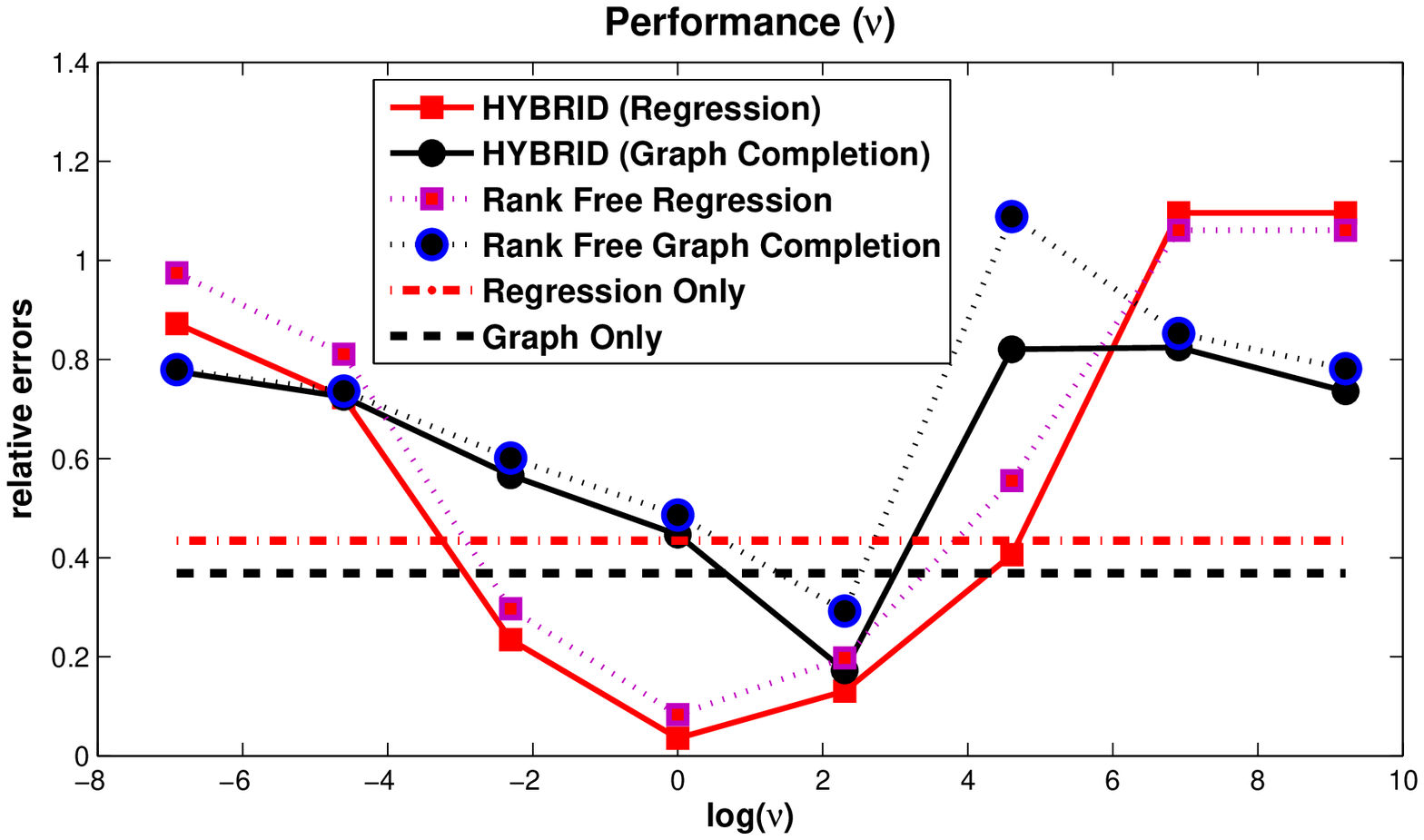}\\
     \includegraphics[height=4.5cm]{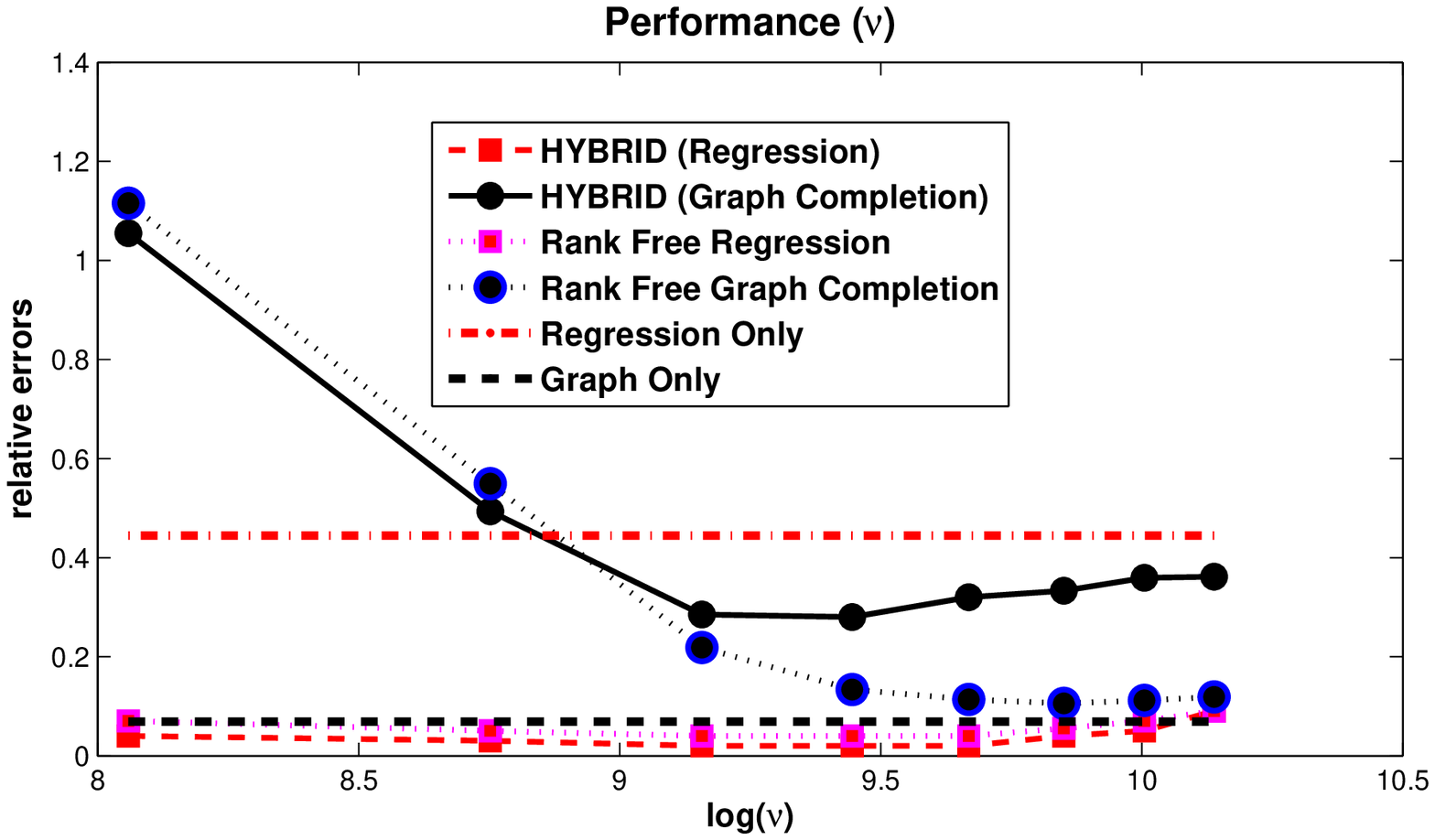}

\caption{Performance on real data for different values of $\nu$, where we have set $\tau = \nu \mu_\text{CV}$ for a value of $\mu_\text{CV}$ fixed in advance by cros-validation. Book categories sales volumes and cross-selling on the top and best-selling items on the bottom. }
\label{fig:mkt}
\end{figure}

We implemented a first-order gradient descent algorithm with
projections on the convex set $\mathcal{E}$ at each iteration.  In
Figure \ref{fig:cv} we show how cross validation can be done first on
the two standard objectives of regression and graph completion (matrix
denoising) taken separately for setting the values of $\mu =
\frac{\tau}{\nu}$ and $\kappa$ efficiently. This reduces the number of
smoothing parameters, and thanks to experiments on synthetic data we
studied the dependence of optimal regularization parameters $\kappa,\tau,\nu,\lambda$ on the model
parameters $r,\sigma,\epsilon$. We illustrate in Figure \ref{fig:desc} the efficiency of
our algorithm for minimizing different objective terms despite
competing effects at the starting point of optimization. We 
emphasize the decrease of validation error as a function of the number of iterations. 
Finally, in Figure \ref{fig:mkt}, we report results of experiments on marketing data
sets, which show that, with appropriately chosen hyperparameters, our approach outperforms 
standard competing baselines. We also
observe that on the top-selling items data set, the nuclear norm term
does not add to prediction accuracy compared to the rank-free
$\tau=0$ case. Further study is needed to better understand whether
this effect is due to the possible presense of high noise or to the higher efficiency of
predicted features in predicting the future graph.

\begin{center}
\begin{table}
{\footnotesize
\begin{tabular}{l | c c || c c || c c}
Data & \multicolumn{2}{c||}{Marketing(BS)} & \multicolumn{2}{c||}{Marketing(cat.)} & \multicolumn{2}{c}{Synthetic} \\
\hline
&&&&\\
Method  $\backslash$ Res.        & Feature & Graph & Feature & Graph & Feature & Graph \\
&&&&\\
\hline
Hybrid  & 0.06 & 0.33 & 0.08  & 0.31  &  0.13 $\pm$ .002 & 0.21$ \pm$ .003 \\
$\lambda=0$ : \cite{Richard10}  & 0.12 & 0.32 & 0.09 & 0.34   &  0.16 $\pm$ .003 & 0.22$ \pm$ .004 \\
$\tau = 0$ : Rank-free  & 0.08 & 0.39 & 0.06 & 0.24  & 0.19 $\pm$ .008 & 0.24 $\pm$ .01 \\
Regression Only & -& 0.42& -& 0.17& - &  0.23 $\pm$ .003 \\
Graph Only & 0.40&  - & .31&  - &    0.21 $\pm$ .01 &  - \\
\hline

\end{tabular}
\caption{Comparison of our method with other methods on real marketing data and synthetic data sets. Hyper-parameters set after a cross validation step on past data. Synthetic data results are reported with bootstrap standard type errors computed after 50 replications.  Feature prediction relative residuals $\frac{\|X_{T+1}-f(\Phi_T)\|_2}{\|X_{T+1}\|_2}$ and  Graph prediction relative residuals $\frac{\|A_{T+1}-S\|_F}{\|A_{T+1}\|_F}$ reported.}
}
\label{tab:res}
\end{table}
\end{center}



\section{Conclusion}

We have studied the benefits of joint regularization for simultaneously predicting 
time series which are related, as well as the graph representing these
relations. The main goal has been to investigate the feasibility as well
as the advantages of such a simultaneous estimation procedure.
Although the corresponding optimization problem is not convex, we
identified a convexity domain which in some cases will contain the optimal
solution. The experimental results indicate that exploring such joint
regularization and learning problems can greatly improve the
predictive performance of these methods. Clearly, an important
question is to determine conditions under which the global optimum is attained within
the convexity domain of the optimization problem \eqref{eq:optim}. 
A key contribution of this work is to show the
potential of this hybrid approach, as well as to propose a practical
algorithm for solving the
learning problem. Moreover, given the promising empirical results, 
future improvements in the optimization methodology may lead to
even further improvements in terms of predictive performance.

\bibliographystyle{plain}
\phantomsection\addcontentsline{toc}{section}{\refname}\bibliography{graphlink3}

\end{document}